\pdfoutput=1

\documentclass[11pt]{article}

\usepackage[final]{acl}

\usepackage{times}
\usepackage{latexsym}

\usepackage[T1]{fontenc}

\usepackage[utf8]{inputenc}

\usepackage{microtype}

\usepackage{inconsolata}

\usepackage{graphicx}
\usepackage{amsmath}
\usepackage{amssymb}
\usepackage{mathtools}
\usepackage{amsthm}
\usepackage{hyperref}
\usepackage{amssymb}
\usepackage{multirow}
\usepackage{xcolor}
\usepackage{colortbl}
\usepackage{booktabs}

\usepackage{algorithm}
\usepackage{algorithmicx}
\usepackage{algpseudocode}
\usepackage{verbatim}

\definecolor{marron}{HTML}{CC88B0}

\title{\centering ToolHop: A Query-Driven Benchmark for Evaluating\\ Large Language Models in Multi-Hop Tool Use}

\author{
    \bf{\normalsize
    Junjie Ye$^{1,2}$, \ \ Zhengyin Du$^{2}$, \ \ Xuesong Yao$^{2}$, \ \ Weijian Lin$^{2}$,}\\
    \bf{\normalsize Yufei Xu$^{2}$, \ \ Zehui Chen$^{2}$, \ \ Zaiyuan Wang$^{2}$, \ \ Sining Zhu$^{2}$, \ \ Zhiheng Xi$^{1}$,}\\
    \bf{\normalsize Siyu Yuan$^{1}$, \ \ Tao Gui$^{1,3}$\thanks{Corresponding authors.}, \ \ Qi Zhang$^{1,3*}$, \ \ Xuanjing Huang$^{1,3*}$}, \ \ Jiecao Chen$^{2}$ \\ 
  {$^1$ \normalsize Fudan University \ \ }
  {$^2$ \normalsize ByteDance} \\
  {$^3$ \normalsize Shanghai Collaborative Innovation Center of Intelligent Visual Computing} \\
  \texttt{\normalsize jjye23@m.fudan.edu.cn, \{qz, tgui\}@fudan.edu.cn} \\
  }

\begin{document}
\maketitle
\begin{abstract}
Effective evaluation of multi-hop tool use is critical for analyzing the understanding, reasoning, and function-calling capabilities of large language models (LLMs). However, progress has been hindered by a lack of reliable evaluation datasets. To address this, we present~\emph{ToolHop}, a dataset comprising 995 user queries and 3,912 associated tools, specifically designed for rigorous evaluation of multi-hop tool use. ToolHop ensures diverse queries, meaningful interdependencies, locally executable tools, detailed feedback, and verifiable answers through a novel query-driven data construction approach that includes tool creation, document refinement, and code generation.
We evaluate 14 LLMs across five model families (i.e., LLaMA3.1, Qwen2.5, Gemini1.5, Claude3.5, and GPT), uncovering significant challenges in handling multi-hop tool-use scenarios. The leading model, GPT-4o, achieves an accuracy of 49.04\%, underscoring substantial room for improvement. Further analysis reveals variations in tool-use strategies for various families, offering actionable insights to guide the development of more effective approaches. Code and data can be found in~\url{https://huggingface.co/datasets/bytedance-research/ToolHop}.

\end{abstract}

\section{Introduction}

\begin{figure}
    \centering
    \includegraphics[width=\linewidth]{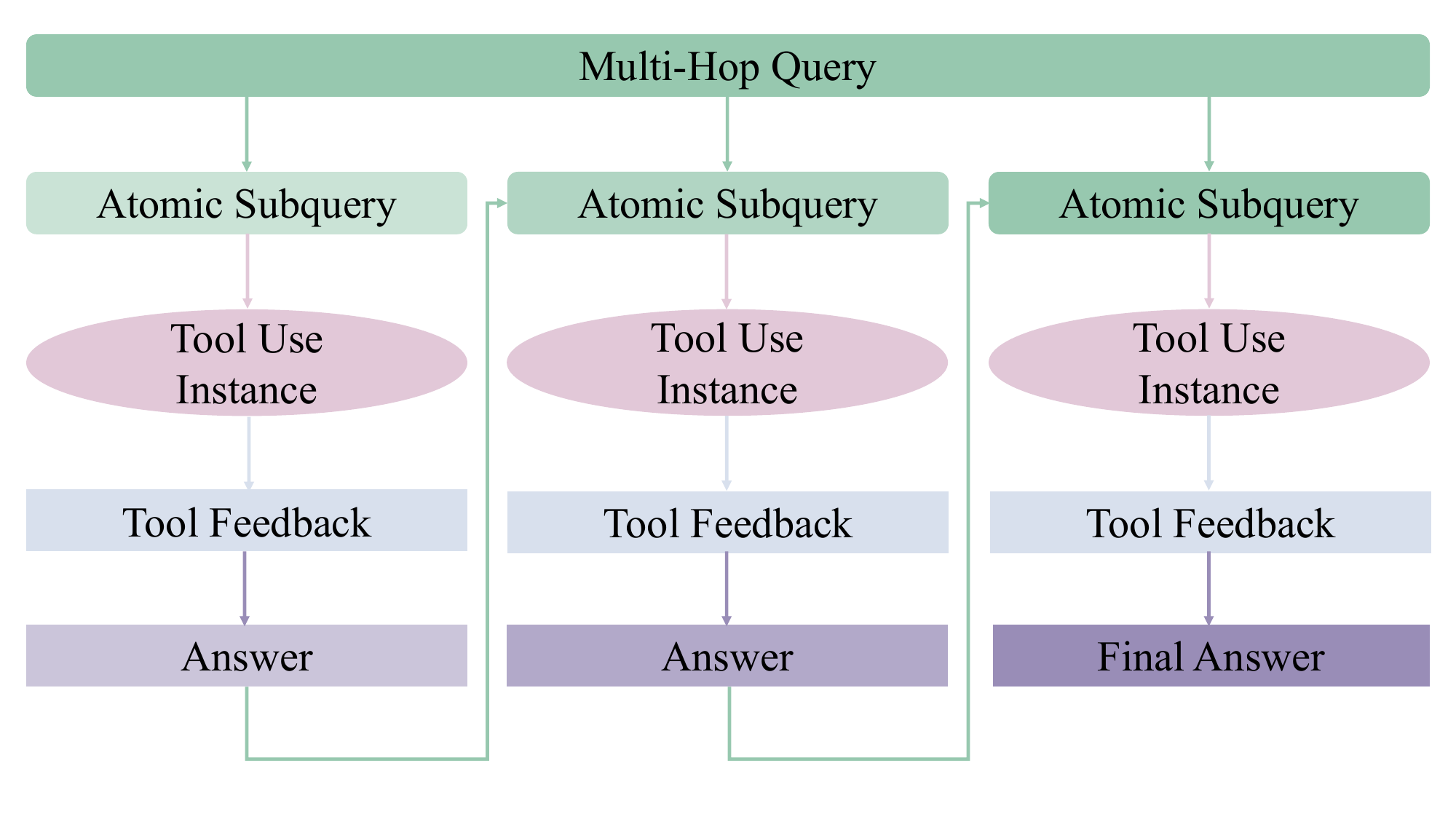}
    \caption{An illustration of multi-hop tool use. The process entails decomposing a complex multi-hop query into multiple atomic sub-queries, sequentially invoking the appropriate tools, retrieving results from the tool feedback, and iterating until the final answer is derived. This demonstrates the integration of comprehension, reasoning, and function-calling capabilities.}
    \label{fig:example}
\end{figure}

\begin{figure*}
    \centering
    \includegraphics[width=\linewidth]{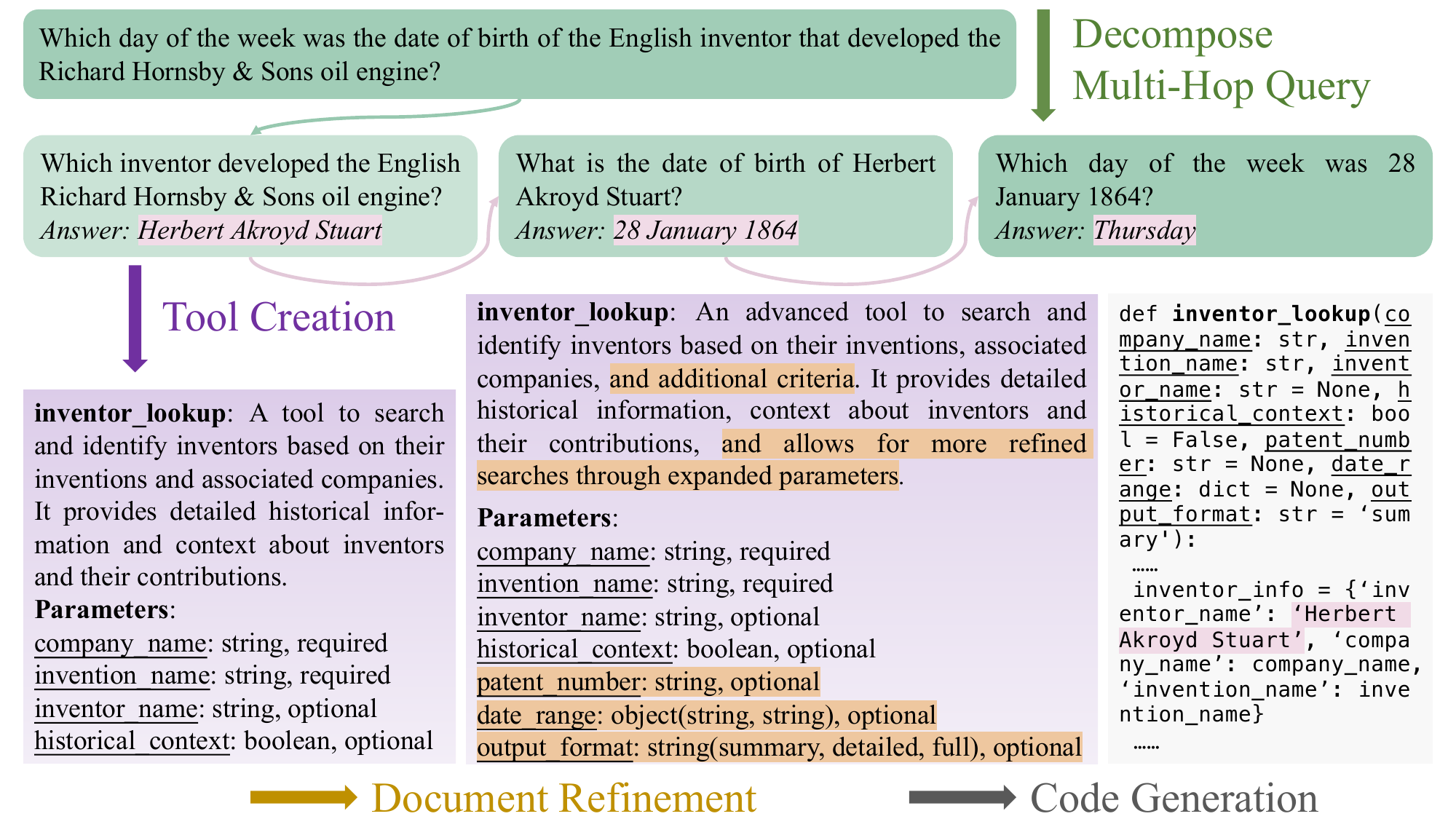}
    \caption{An illustration of our proposed query-driven data construction scheme, comprising three key processes: tool creation, document refinement, and code generation. This approach incrementally develops detailed tool document and code implementation for each atomic subquery within a multi-hop query.}
    \label{fig:scheme}
\end{figure*}

The task of multi-hop tool use presents a significant challenge for large language models (LLMs)~\cite{GPT-4, LLaMA-2, Qwen}. As illustrated in Figure~\ref{fig:example}, it requires LLMs to incrementally decompose a complex multi-hop query into atomic subqueries, invoke the appropriate tools, and iteratively retrieve results from the tool feedback until the final answer is reached. This process demands advanced capabilities such as comprehension, reasoning, and function-calling~\cite{tool-learning, tool-survey}, making the evaluation of multi-hop tool use essential for assessing these skills. Furthermore, such evaluations are pivotal for advancing LLMs toward generalized intelligence~\cite{agent-survey}.

Existing studies have made progress in evaluating tool use of LLMs. Some focus on evaluating single-step tool use in simulation environments, requiring manual calibration of correct tool-call results~\cite{TEval, ToolSword, RoTBench}. Others examine the process of tool use, leveraging advanced models like GPT-4 to go beyond single-step evaluations and providing some valuable insights~\cite{Toolllm, metatool, ToolEyes}.

However, these works still fall short of offering a reliable evaluation of multi-hop tool use. Specifically, a key limitation of prior work lies in their reliance on tool-driven data construction methods, where a collection of tools is gathered and queries are simulated for them~\cite{Toolalpaca, Sealtools, ToolACE}. This approach fails to ensure that the collected tools are interdependent or that the queries involve genuine multi-hop reasoning. Furthermore, the absence of verifiable answers forces these studies to depend on process analysis using models, introducing model bias and evaluation errors~\cite{eval-survey, fairness}.

To address these challenges, we introduce~\emph{ToolHop}, a novel dataset specifically designed to evaluate LLMs' multi-hop tool use capabilities. ToolHop comprises 995 multi-hop queries and 3,912 locally executable tools, constructed using a query-driven data construction scheme. This methodology involves tool creation, document refinement, and code generation, which can expand a single multi-hop query into a comprehensive multi-hop tool use test case. An analysis of ToolHop demonstrates its effectiveness in accommodating diverse queries, ensuring meaningful interdependencies, supporting locally executable tools, and delivering detailed feedback alongside verifiable answers. This design rigorously evaluates LLMs' multi-hop tool use capabilities.

We evaluate ToolHop on 14 LLMs from five different families (i.e., LLaMA3.1~\cite{LLaMA3.1}, Qwen2.5~\cite{Qwen2.5}, Gemini1.5~\cite{Gemini1.5}, Claude3.5~\cite{Claude}, and GPT~\cite{GPT-4}). Our results reveal that while tools significantly improve model performance, even the top-performing model, GPT-4, achieves only 49.04\% accuracy in multi-hop tool use, highlighting considerable room for improvement. Further studies reveal that different model families exhibit distinct patterns in tool use, leading to varied outcomes. For instance, the Qwen2.5~\cite{Qwen2.5} family of models tends to emphasize parallel calls, which results in hallucinations, while the GPT family leverages tool feedback to improve their performance in tool usage. These insights provide valuable guidance for developing more effective methods.

Our contributions are as follows: 
\begin{itemize} 
\item We introduce ToolHop, a test set of 995 multi-hop queries with 3,912 locally executable tools, designed to assess LLMs’ ability to use tools in multi-hop scenarios. It ensures diverse queries, meaningful interdependencies, locally executable tools, detailed feedback, and verifiable answers.
\item We propose a novel query-driven data construction process that can expand queries into multi-hop tool use data via tool creation, document refinement, and code generation.
\item We provide a comprehensive evaluation of 14 LLMs, identifying significant limitations in current tool-use capabilities and offering insights for future improvements.
\end{itemize}

\section{ToolHop}

In this section, we introduce ToolHop in detail. Specifically, we first provide a formal definition of multi-hop tool use (Section~\ref{sec:formulation}), then explain our proposed query-driven data construction scheme (Section~\ref{sec:construction}), and finally analyze the quality of the ToolHop dataset (Section~\ref{sec:data-analysis}).

\subsection{Task Formulation}
\label{sec:formulation} 
Given a multi-hop query \( q \), which can be decomposed into subqueries \( q_1, q_2, ..., q_l \); and a collection of tools \( \mathbb{T} = (t_1, t_2, \dots, t_l) \), where each tool \( t_i \) is defined by a document \( \text{doc}_i \) and a code implementation \( \text{fun}_i \), the description document \( \text{doc}_i \) includes the tool name \( n_i \), a function description \( d_i \), and the corresponding parameters \( p_i = (p_i^1, p_i^2, \dots, p_i^k) \).Each parameter \( p_i^j \) is characterized by its name \( np_i^j \), a description \( dp_i^j \), its type \( tp_i^j \), and whether it is required \( rp_i^j \).
The goal of multi-hop tool use is for the model \( \mathcal{M} \) to utilize the information in \( \mathbb{T} \) to sequentially invoke the appropriate tools \( t_1, t_2, ..., t_l \), where each tool \( t_i \) is used to solve subquery \( q_i \) and produce an intermediate answer \( a_i \). The output \( a_i \) then serves as one of the inputs to the next tool \( t_{i+1} \), enforcing the need for the model to correctly understand complex dependencies between tools and accurately decompose the original query. The final answer \( a \) is obtained after executing the full sequence of tool calls.
A visual illustration of this process is provided in Figure~\ref{fig:example}.

\subsection{Query-Driven Data Construction}
\label{sec:construction}

As illustrated in Figure~\ref{fig:scheme}, we propose a novel query-driven data construction scheme that departs from traditional tool-driven approaches. This scheme comprises three key stages that involves tool creation, document refinement, and code generation. Given a multi-hop user query \( q \), the scheme extends \( q \) to produce a sequence of corresponding tool documents \( \text{doc}_{i..l} \) and their associated code implementations \( \text{fun}_{i..l} \).

\paragraph{Tool Creation}

The query-driven data construction begins with the multi-hop user query \( q \), which serves as the foundation for building dynamic tools. The tool creation process accepts \( q \) and generates a preliminary set of tool documents \( \text{doc}'_{1..l} \). These documents are designed to be both relevant to \( q \) and interdependent. 

To achieve this, \( q \) is decomposed into a sequence of atomic subqueries \( q_1, q_2, \dots, q_l \), where each subquery \( q_i \) depends on resolving the preceding ones (i.e., \( q_{i-1} \)). For each \( q_i \), a preliminary document \( \text{doc}'_i \) is created. . These documents not only capture the input-output logic of \(q_i\), but are also structured to generalize to similar queries. By maintaining backward and forward dependencies between documents, this approach ensures both modularity and cohesion, simplifying the tool creation process.

\paragraph{Document Refinement}

The initial tool documents \( \text{doc}'_i \), derived directly from atomic queries, are typically rudimentary due to the limited information in \( q_i \). The document refinement process transforms \( \text{doc}'_i \) into a more comprehensive document \( \text{doc}_i \), designed to better support the evaluation of models in complex multi-hop scenarios.

This process involves two key aspects. On the one hand, the tool's functionality is expanded by introducing features such as result filtering and customizable formats, all while maintaining compatibility with the original functionality. On the other hand, the number of parameters is increased, and their types are optimized. For instance, parameters initially represented as simple strings are replaced with structured types such as arrays or objects, enabling the tools to handle more complex inputs. These refinements ensure that the resulting tool documents are robust, versatile, and capable of addressing intricate cases.

\begin{figure}[!t]
    \centering
    \includegraphics[width=\linewidth]{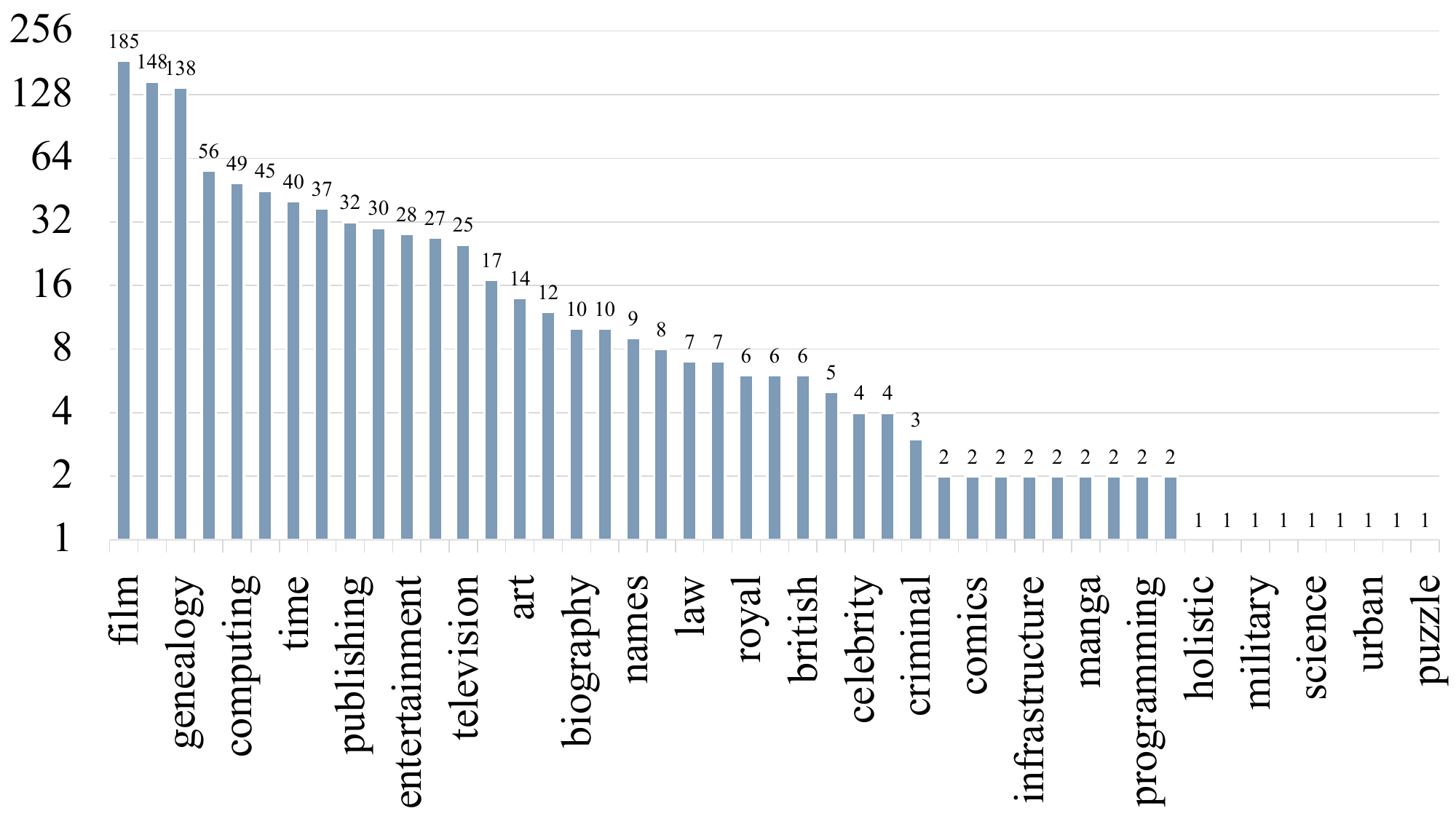}
    \caption{Distribution of user queries across 47 domains in the ToolHop dataset.}
    \label{fig:domain}
\end{figure} 

\begin{table}[!t]
    \centering
    \resizebox{0.85\linewidth}{!}
    {
    \begin{tabular}{lccccc}
    \toprule
        \textbf{\# Tools} & Three & Four & Five & Six & Seven  \\ \midrule
        \textbf{\# Data} & 428 & 353 & 136 & 10 & 68 \\
        \bottomrule
    \end{tabular}
    }
    \caption{Distribution of the number of tools required to solve each query in ToolHop.}
    \label{tab:tools}
\end{table}

\begin{figure}[!t]
    \centering
    \includegraphics[width=\linewidth]{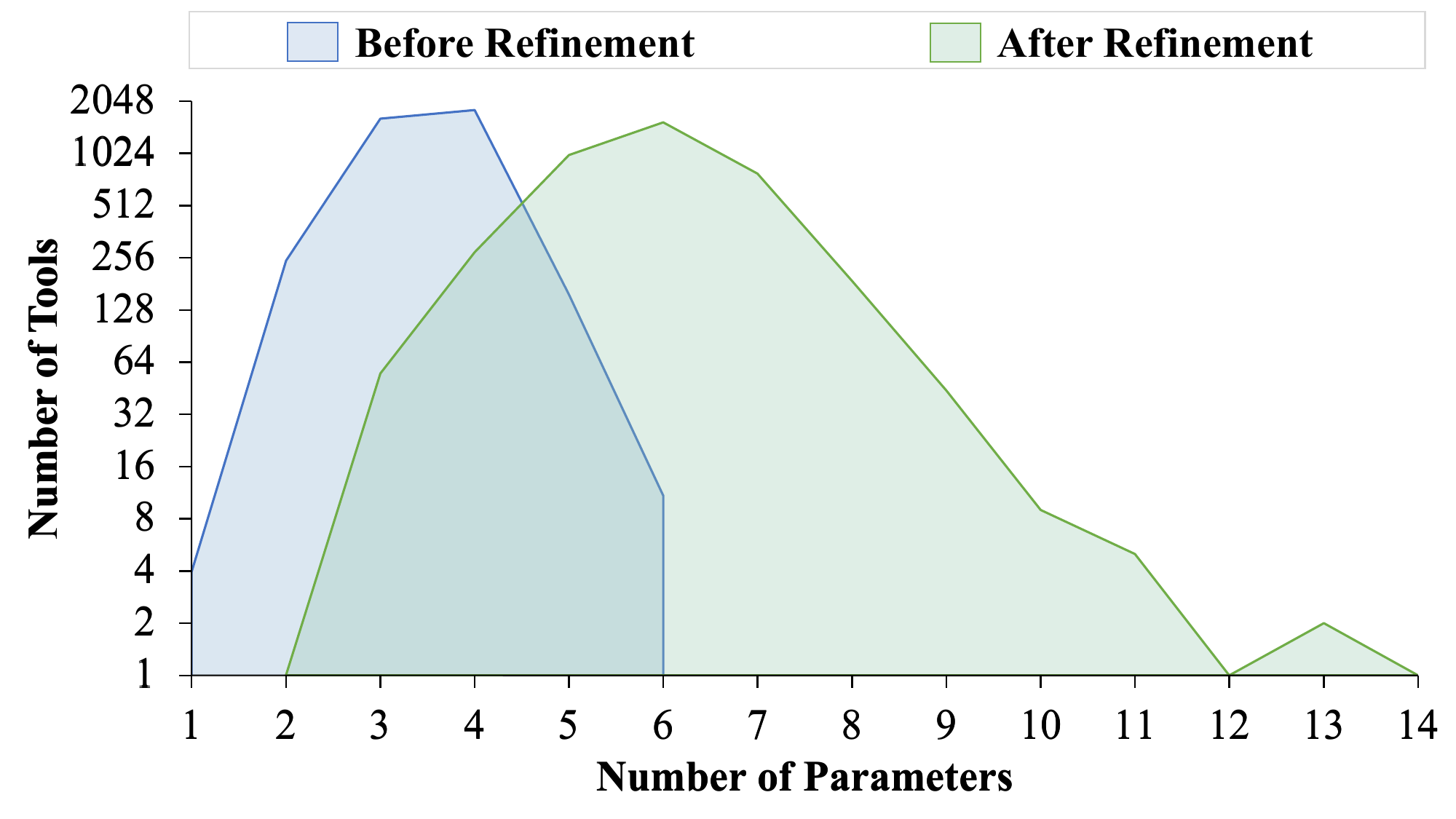}
    \caption{Distribution of the number of tool parameters before and after document refinement.}
    \label{fig:param_num}
\end{figure}

\paragraph{Code Generation}

Once refined tool documents \( \text{doc}_i \) are complete, the code generation process produces corresponding locally executable functions \( \text{fun}_i \). These functions allow external invocation of tools, enabling seamless multi-turn interactions between the model and tools.

Code generation systematically maps document information to code. For instance, the tool name in $\text{doc}_i$ is converted into the function name, while parameter specifications are used to define the function signature. To ensure the correctness of \( \text{fun}_i \), the atomic query \( q_i \) and its answer \( a_i \) are included as inputs, requiring  the function to return \( a_i \) when executed with \(q_i\). Additionally, a robust exception-handling mechanism is implemented, enabling tools to provide informative error messages for invalid inputs while maintaining normal operation. Moreover, the generated code is verified to ensure it functions as intended.

\paragraph{Dataset Construction}
To effectively implement our approach, we draw on queries from the MoreHopQA dataset~\cite{morehopqa}, which consists of multi-hop questions that can be decomposed into at least three atomic queries with answers. Using this foundation, we generate 995 user queries and 3,912 corresponding locally executable tools, which collectively form the ToolHop dataset.\footnote{Examples of generated documents and code implementations are provided in Appendix~\ref{sec:example-document} and Appendix~\ref{sec:example-code}.}

\subsection{Dataset Analysis}  
\label{sec:data-analysis}  

To ensure that the ToolHop dataset rigorously evaluates the multi-hop tool-use capabilities of LLMs, we conduct a comprehensive analysis across five critical dimensions. This analysis validates ToolHop’s ability to represent diverse and challenging multi-hop tool-use scenarios.\footnote{A comparison between ToolHop and existing benchmarks related to tool use can be found in Appendix~\ref{sec:comparison}.}

\begin{figure}[!t]
    \centering
    \includegraphics[width=\linewidth]{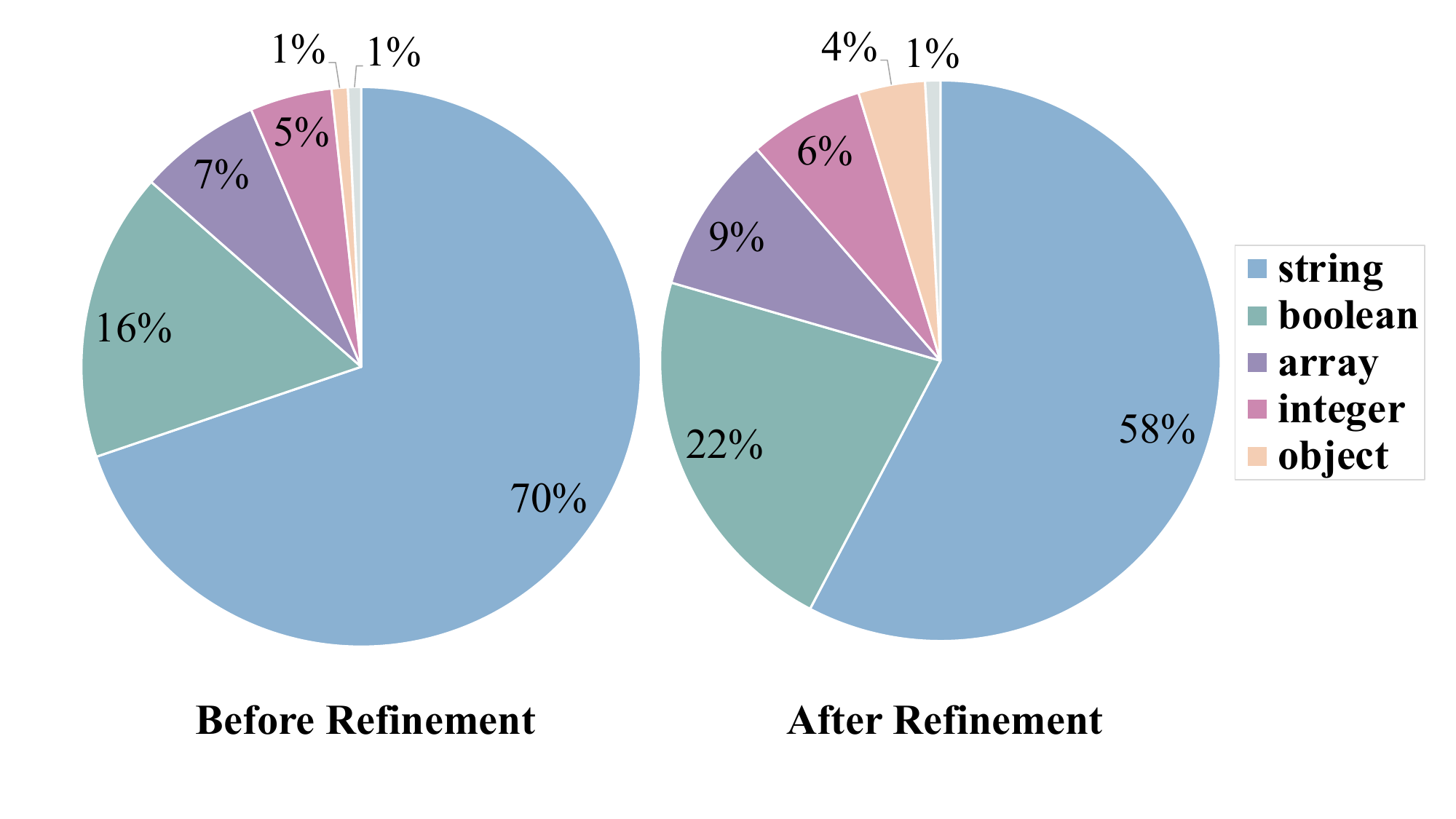}
    \caption{Distribution of tool parameter types before and after document refinement.}
    \label{fig:param_type}
\end{figure}

\paragraph{Diverse Queries}

Real-world user needs vary widely, requiring an effective LLM to flexibly utilize tools to address queries spanning multiple domains. To evaluate such capabilities, a suitable dataset must encompass queries from a broad range of topics. ToolHop is explicitly designed to prioritize diversity in its multi-hop queries, reflecting real-world scenarios.

To verify this diversity, we use GPT-4o to categorize all queries in ToolHop into distinct domains. Similar categories are merged to ensure clarity and independence. As shown in Figure~\ref{fig:domain}, the categorization reveals that ToolHop spans 47 unique domains, including topics such as movies and television, academic subjects, and family relationships. This broad coverage ensures that ToolHop effectively evaluates LLM performance across diverse query types, enhancing its representativeness and practical applicability for real-world tool-use scenarios.

\paragraph{Meaningful Interdependencies}
Previous evaluation for tool use~\cite{RestGPT, GPT4tool, RoTBench, NesTools} typically assemble tools from disparate sources and then generate user queries for them. However, these approaches fail to account for interdependencies between tools, often producing queries that inadequately represent multi-hop reasoning. To address this limitation, ToolHop employs a novel query-driven framework. It begins by formulating multi-hop queries and subsequently constructs the required tools based on each atomic query. This approach inherently preserves the multi-hop structure of queries and enforces meaningful interdependencies between tools.

To validate the effectiveness of this approach, we analyze the distribution of tools associated with each query in ToolHop. As shown in Table~\ref{tab:tools}, the number of tools required per query ranges from three to seven, which corresponds to the minimum number of reasoning hops required to resolve the queries, emphasizing the importance of multi-hop reasoning. This distribution underscores the complexity of queries handled by ToolHop and its capability to support scalable multi-hop tool use.

\paragraph{Locally Executable Tools}

Tools are a core component of the tool use task. ToolHop includes 3,912 locally deployable and directly executable tools, enabling zero-cost invocation and seamless interaction by LLMs. To better align the constructed tools with the diverse requirements of real-world applications, we enhance their complexity through a document refinement process.

Figure~\ref{fig:param_num} shows that the average number of parameters per tool increased from 3.49 to 5.91 after refinement. This reflects an intentional shift toward more expressive tools, which better capture the complexity of real-world tasks. Concurrently, Figure~\ref{fig:param_type} illustrates a 12\% reduction in simple string parameters, replaced by more structured types such as arrays, booleans, and objects, which enable richer and more precise tool interactions. Table~\ref{tab:param_num} and Table~\ref{tab:param_type} further demonstrate that the refinement process preserves the number and types of required parameters while increasing the diversity of optional parameters.

\begin{table}[!t]
    \centering
    \resizebox{0.85\linewidth}{!}
    {
    \begin{tabular}{lcccccc}
    \toprule
      \textbf{Refinement}   & \textbf{Zero} & \textbf{One} & \textbf{Two} & \textbf{Three} & \textbf{Four}  \\ \midrule
        Before & 2 & 2433 & 1250 & 202 & 25 \\
        After & 2 & 2490 & 1198 & 200 & 22 \\
        \bottomrule
    \end{tabular}
    }
    \caption{Distribution of the number of required parameters before and after document refinement.}
    \label{tab:param_num}
\end{table}

\begin{table}[!t]
    \centering
    \resizebox{\linewidth}{!}
    {
    \begin{tabular}{lccccccc}
    \toprule
      \textbf{Refinement}   & \textbf{string} & \textbf{boolean} & \textbf{array} & \textbf{integer} & \textbf{object} & \textbf{number}  \\ \midrule
        Before & 4758 & 2 & 404 & 333 & 24 & 114 \\
        After & 4473 & 2 & 755 & 241 & 44 & 102 \\
        \bottomrule
    \end{tabular}
    }
    \caption{Distribution of required tool parameter types before and after refinement.}
    \label{tab:param_type}
\end{table}

\paragraph{Detailed Feedback}

Effective multi-turn interaction between LLMs and tools requires not only correct outputs for valid inputs but also meaningful error messages for invalid ones. Our approach incorporates two key strategies to address this need.

On the one hand, we include atomic queries and their corresponding answers as part of the input during code generation, ensuring tools reliably produce correct outputs for solvable problems. On the other hand, we integrate robust exception-handling mechanisms into the generated code. Since the tools are locally executable, we can validate LLM-generated call instances using a compiler, providing detailed error reports and feedback to guide subsequent interactions.

\paragraph{Verifiable Answers} 

A key limitation of earlier tool-driven datasets is the absence of predetermined answers, which makes validation difficult. ToolHop overcomes this issue by predefining both queries and answers, enabling straightforward comparison with model outputs.

\begin{figure}[!t]
    \centering
    \includegraphics[width=\linewidth]{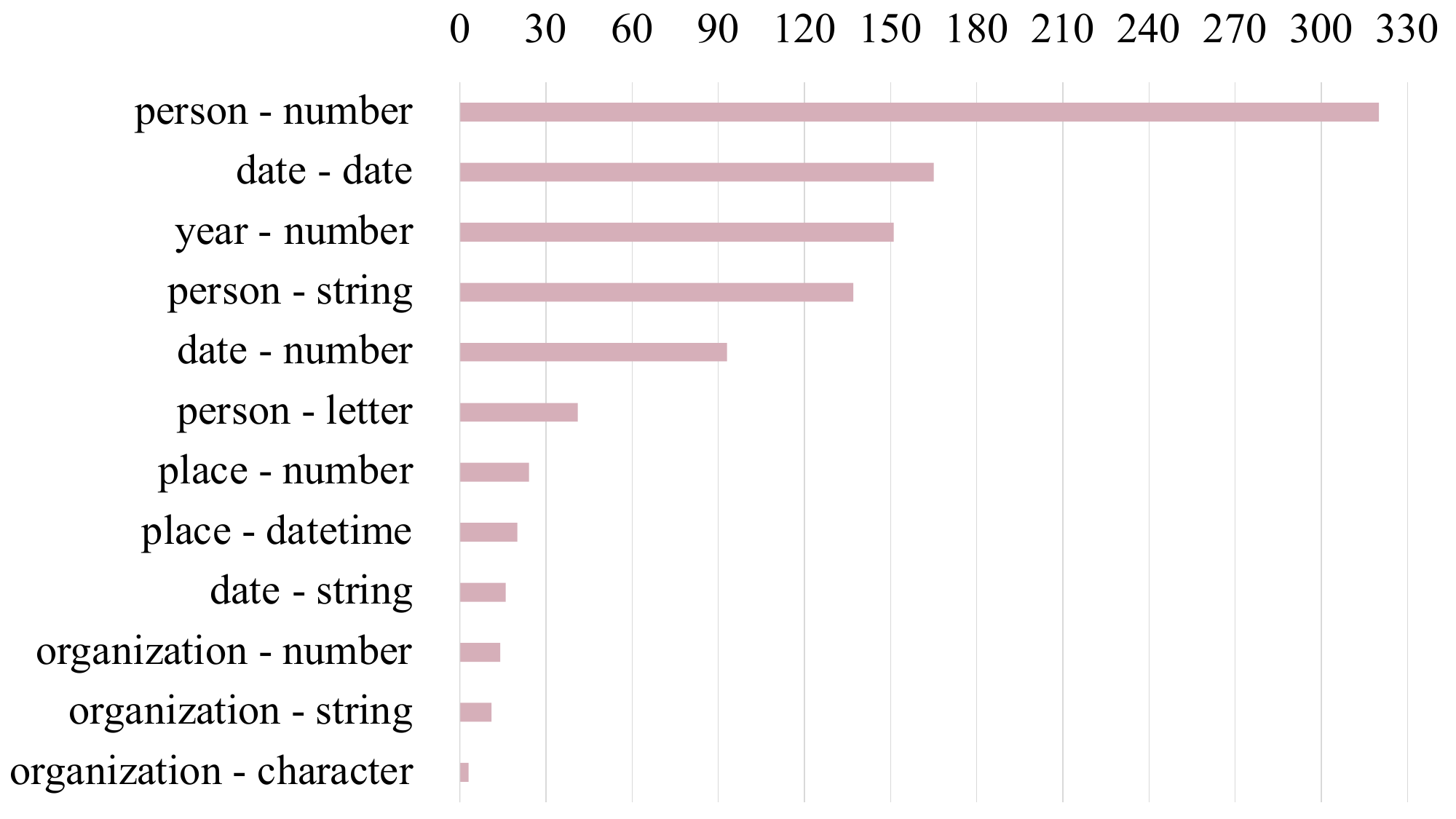}
    \caption{Distribution of answer types for the second atomic subquery and final answers in ToolHop.}
    \label{fig:answer}
\end{figure}

To ensure verifiability, we analyze the answer types for the second atomic subquery and the final query, which is presented in Figure~\ref{fig:answer}. The result demonstrates that ToolHop supports diverse and flexible answer types while standardizing final answers into objective entities. This design simplifies validation, enhances robustness, and enables consistent performance evaluation.

\section{Experimental Setup}  
We use ToolHop to evaluate the representative families of LLMs for multi-hop tool use. In this section, we introduce the families of LLMs evaluated (Section~\ref{sec:models}) and describe the implementation details of our experiments (Section~\ref{sec:details}).  

\begin{table*}[!t]
    \centering
    \resizebox{\linewidth}{!}
    {
    \begin{tabular}{lllccccc}
    \toprule
    \multirow{2}*{\textbf{Source}} & \multirow{2}*{\textbf{Family}} & \multirow{2}*{\textbf{Version}} & \multicolumn{3}{c}{\textbf{Answer Correctness} ($\uparrow$)} & \multicolumn{2}{c}{\textbf{Invocation Error} ($\downarrow$)} \\ \cmidrule(lr){4-6} \cmidrule(lr){7-8}
   & & & \textbf{Direct} & \textbf{Mandatory} & \textbf{Free} & \textbf{Query} & \textbf{Instance} \\ \midrule 
    & & \textit{Avg.} & \textit{19.83} & \textit{32.12} & \textit{32.84} & \textit{18.72} & \textit{8.68} \\ \midrule 
      \multirow{6}*{Open-Source} & \multirow{2}*{LLaMA3.1} & Instruct-8B & \cellcolor{marron!18}13.17 & \cellcolor{marron!60}12.76 & \cellcolor{marron!57}13.47  & \cellcolor{marron!69}41.61 & \cellcolor{marron!39}21.10\\
      & &Instruct-70B & \cellcolor{marron!3}18.79 & \cellcolor{marron!39}19.10 & \cellcolor{marron!60}12.76 & \cellcolor{marron!51}35.08 & \cellcolor{marron!18}14.24 \\ \cmidrule{2-8} 
      & \multirow{4}*{Qwen2.5} & Instruct-7B & \cellcolor{marron!24}11.46 & \cellcolor{marron!69}9.85 & \cellcolor{marron!48}16.18 & \cellcolor{marron!30}28.84 & \cellcolor{teal!3}7.09 \\
      & & Instruct-14B & \cellcolor{marron!6}17.39 & \cellcolor{marron!18}26.38 & \cellcolor{marron!18}26.13 & \cellcolor{teal!9}15.78 & \cellcolor{teal!6}6.82 \\
      & & Instruct-32B & \cellcolor{teal!3}20.00 & \cellcolor{marron!21}25.03 & \cellcolor{marron!30}22.61 & \cellcolor{teal!18}12.46 & \cellcolor{teal!15}3.46 \\
      & & Instruct-72B & \cellcolor{marron!6}17.89 & \cellcolor{teal!39}45.43 & \cellcolor{teal!18}38.29 & \cellcolor{teal!15}13.27 & \cellcolor{teal!12}4.93 \\ \midrule 
      \multirow{8}*{Closed-Source} & \multirow{2}*{Gemini1.5} & flash-002 & \cellcolor{marron!3}18.59 & \cellcolor{marron!9}29.35 & \cellcolor{marron!1}32.76 & \cellcolor{teal!15}13.59 & \cellcolor{teal!6}6.69 \\
      & & pro-002 & \cellcolor{marron!3}18.89 & \cellcolor{marron!3}31.16 & \cellcolor{teal!3}33.07 & \cellcolor{teal!12}14.57 & \cellcolor{teal!6}6.61 \\ \cmidrule{2-8} 
      & \multirow{2}*{Claude3.5} & Haiku & \cellcolor{teal!51}36.08 & \cellcolor{teal!18}38.09 & \cellcolor{teal!36}44.72 & \cellcolor{marron!15}23.48 & \cellcolor{marron!21}15.81 \\
      & & Sonnet & \cellcolor{teal!24}27.14 & \cellcolor{teal!21}39.90 & \cellcolor{teal!39}45.23 & \cellcolor{marron!3}19.60 & \cellcolor{marron!21}15.83 \\ \cmidrule{2-8} 
      & \multirow{4}*{GPT} & 3.5-Turbo & \cellcolor{marron!6}17.09 & \cellcolor{teal!9}35.38 & \cellcolor{teal!12}36.58 & \cellcolor{teal!21}11.76 & \cellcolor{teal!6}6.03 \\
      & & 4o-mini & \cellcolor{marron!1}19.40 & \cellcolor{teal!24}40.20 & \cellcolor{teal!33}43.42 & \cellcolor{teal!21}11.66 & \cellcolor{teal!15}3.58 \\
      & & 4-Turbo & \cellcolor{marron!3}18.59 & \cellcolor{teal!45}47.94 & \cellcolor{teal!42}46.83 & \cellcolor{teal!24}10.95 & \cellcolor{teal!12}4.97 \\
      & & 4o & \cellcolor{teal!12}23.12 & \cellcolor{teal!51}49.04 & \cellcolor{teal!45}47.74 & \cellcolor{teal!27}9.45 & \cellcolor{teal!12}4.31 \\
      \bottomrule
    \end{tabular}
    }
    \caption{Performance of various LLMs on ToolHop, including answer correctness and invocation error. `Direct,' `Mandatory,' and `Free' denote the direct answer, mandatory tool use, and free choice scenarios, respectively. `Query' and `Instance' refer to the percentage of queries and tool invocation instances with errors, respectively. `\textit{Avg.}' represents the average across all LLMs. Values above the average are highlighted in \colorbox{teal}{teal}, and those below are highlighted in \colorbox{marron}{maroon}, with darker shades indicating larger deviations.}
    \label{tab:result}
\end{table*}

\subsection{Models}  
\label{sec:models}  
We use ToolHop to evaluate 14 LLMs from five families, including \textbf{LLaMA3.1-Instruct-8B} and \textbf{LLaMA3.1-Instruct-70B} from the LLaMA3.1 family, \textbf{Qwen2.5-Instruct-7B}, \textbf{Qwen2.5-Instruct-14B}, \textbf{Qwen2.5-Instruct-32B}, and \textbf{Qwen2.5-Instruct-72B} from the Qwen2.5 fmaily, \textbf{Gemini1.5-flash-002} and \textbf{Gemini1.5-pro-002} from the Gemini1.5 family, textbf{Claude3.5-Haiku} and \textbf{Claude3.5-Sonnet} from the Claude3.5 family, and \textbf{GPT-3.5-Turbo}, \textbf{GPT-4o-mini}, \textbf{GPT-4-Turbo}, and \textbf{GPT-4o} from the GPT family.\footnote{More details can be found in Appendix~\ref{sec:detail-model}.}

\subsection{Implementation Details}  
\label{sec:details}  
In the data construction stage, we use GPT-4o to assist with processing.\footnote{Prompts are detailed in Appendix~\ref{sec:prompt-data}.} 
For evaluation, open-source LLMs are tested using their chat templates with greedy decoding, while closed-source LLMs are evaluated via their APIs with a temperature setting of 0.
To ensure consistency across evaluations, all tools are implemented through the models' function call interfaces.

\section{Main Results}

In this section, we present the key evaluation dimensions (Section~\ref{sec:aspect}) and observations (Section~\ref{sec:analysis}).

\subsection{Evaluation Dimensions}
\label{sec:aspect}

Evaluating the capabilities of LLMs requires a comprehensive approach that assesses both their ability to provide correct answers and their effectiveness in invoking external tools. We analyze these dimensions through answer correctness and invocation error.

\paragraph{Answer Correctness}
For the accuracy of LLM responses, our query-driven data construction scheme enables direct comparison with predefined standard answers. We consider three evaluation scenarios: the \textbf{direct answer} scenario, where LLMs solve queries independently without external tools; the \textbf{mandatory tool use} scenario, where models are required to use provided tools extensively to maximize their tool-use capabilities; and the \textbf{free choice} scenario, where external tools are available but optional, allowing LLMs to balance independent problem-solving with tool use.

\paragraph{Invocation Error}
In the mandatory tool use scenario, we assess errors made when invoking tools, leveraging detailed feedback for each tool to identify errors. We focus on three types: tool hallucination, where models invoke tools not included in the provided toolset; parameter hallucination, where unprovided parameters are used for a given tool; and parameter missing, where required parameters for a tool are omitted. Errors are quantified from the percentage of queries containing incorrect calls relative to \textbf{total queries}, and the percentage of incorrect tool invocations relative to \textbf{all tool use instances}.

\subsection{Evaluation Observations}
\label{sec:analysis}

From the results presented in Table~\ref{tab:result}, we can make several notable observations.\footnote{Model performance under various inference frameworks can be found in Appendix~\ref{sec:further}.}

\textbf{While LLMs have significantly enhanced their ability to solve complex multi-hop queries with the use of tools, their multi-hop tool use capabilities still leave considerable room for improvement.} Comparing the direct answer scenario (i.e., Direct) versus the mandatory tool use scenario (i.e., Mandatory), we observe that the use of tools increases LLMs' answer correctness by an average of 12.29\%. Notably, the GPT family of models improves its accuracy by an average of 23.59\% through tool use, underscoring how effective tool-use capabilities enhance their performance in solving complex multi-hop problems.  
Despite these improvements, the overall accuracy in the mandatory tool use scenario remains limited. Even the best-performing model, GPT-4o, achieves only 49.04\% answer correctness in this scenario. Furthermore, 9.45\% of queries exhibit hallucinations. The performance of LLaMA3.1-Instruct-8B reveals further challenges, with over 40\% of queries containing invocation errors, underscoring the need for better documentation understanding.

\textbf{The performance of different LLM families indicates that most are optimized for tool use, but they exhibit distinct characteristics when solving multi-hop queries.} In both the mandatory tool use scenario and the free choice scenario (i.e., Free), LLMs generally opt to use tools, with answer correctness in these two conditions differing by only 0.62\%. This indicates that most LLMs are specifically optimized for tool use. However, different LLM families show varying strengths in their tool use. For instance, Qwen2.5-Instruct-72B improves its answer correctness by 27.54\% through tool use, while the Claude3.5 family excels in the direct answer scenario without tool reliance. The underlying reasons for these differences are explored in depth in Section~\ref{sec:case}.

\textbf{Examining the performance of different versions within each LLM family, larger models generally demonstrate better tool use to meet user needs, aligning with the scaling law}~\cite{scale_1, scale_2}\textbf{.} Both open-source and closed-source LLMs show an increase in answer correctness and a decrease in invocation error in the mandatory tool use scenario as model size grows. 
Notably, the correlation between invocation errors and answer correctness is stronger at the query level than at the instance level, suggesting that invocation errors in specific queries significantly impair problem-solving.
Interestingly, this pattern enables the inference of relative model sizes within families. For instance, based on performance patterns, GPT-4o is likely a larger and more advanced version compared to other models in the GPT family.

\section{Further Studies}
\label{sec:case}

From the results in Section~\ref{sec:analysis}, we observe significant variation in the performance across different families of LLMs. To further investigate these differences, we analyze each family in detail and present the following key observations.\footnote{Examples illustrating these observations can be found in Appendix~\ref{sec:example}.}

\textbf{The LLaMA3.1 and Gemini1.5 families perform poorly in multi-hop tool use scenarios compared to other LLMs from the same source, primarily due to their incomplete support for tool use capabilities. } 
In the case of LLaMA3.1, the inability to output both natural language text and tool call instances simultaneously restricts its capacity to perform chain-of-thought (CoT)~\cite{CoT} reasoning during tool use, hampering its understanding and analysis of user intent. On the other hand, the Gemini1.5 family of models lack support for union-type parameters, which prevents them from handling tool lists that include complex parameter structures. This limitation significantly reduces their effectiveness in such scenarios.

\textbf{The enhancement of the Qwen2.5 family with parallel tool calls introduces a trade-off between efficiency and accuracy.} Compared to the LLaMA3.1 family, the Qwen2.5 family has improved its ability to utilize tools, particularly with the addition of parallel invocation, which is intended to increase the problem-solving efficiency. However, in multi-hop tool use scenarios, forcing parallel invocation without first processing the results of previous tool calls leads to hallucinations in parameter value assignments, resulting in incorrect answers. For instance, in the mandatory tool use scenario, the percentage of queries involving parallel tool calls is 70.1\% for Qwen2.5-Instruct-14B and even higher at 75.08\% for Qwen2.5-Instruct-32B, contributing to their relatively poor performance. In contrast, Qwen2.5-Instruct-72B reduces the percentage of parallel calls to just 3.82\%, significantly improving its performance.  

\textbf{The optimization of CoT reasoning in the Claude family of models gives them a distinct advantage in the direct answer scenario.} Even without explicit CoT prompts, the Claude3.5 family of models independently adopt a step-by-step CoT approach to decompose user queries and generate answers. This method significantly improves their accuracy compared to other LLMs in such scenarios. For instance, in the direct answer scenario, Claude3.5-Haiku applies CoT reasoning to 64.92\% of queries, while Claude3.5-Sonnet does so for 8.5\%. Additionally, the Claude3.5 family of models do not fully rely on the answers returned by tools. This allows them to produce correct responses using their own internal knowledge when tool invocations lead to errors. Despite a relatively high tool invocation error rate, this ability explains why overall answer correctness remains high.

\textbf{The GPT family of models demonstrates some ability to correct tool call behavior after an error occurs, but this heavily depends on the level of detail in the feedback provided.} Leveraging our query-driven data construction process, we offer detailed feedback when a tool call fails.  
We calculate the percentage of queries with call errors in the mandatory tool use scenario where the GPT family of models ultimately provide the correct answer.  We compare this to the percentage of correct answers when only minimal feedback is given, such as a simple hint indicating the call failed (e.g., `Failed!').
As shown in Table~\ref{tab:feedback}, the GPT family of models exhibit a significant improvement in performance when detailed feedback is provided, successfully correcting their behavior to arrive at the correct answer. However, when only basic error hints are provided, the correctness of their final answers drops by 20.66\%. This highlights not only the importance of detailed feedback but also the challenges in further enhancing the models' correction capabilities.

Based on these observations, we propose the following recommendations to enhance the model's tool use capabilities in the future:  
1) Develop a robust and adaptable tool-use model that can support a wide range of complex tools;
2) Optimize the model’s parallelism and other capabilities while prioritizing improvements in its understanding of user intent to avoid potential negative effects; 
and 3) Investigate effective strategies for leveraging rich tool feedback to enhance the model’s error correction abilities.  

\begin{table}[!t]
    \centering
    \resizebox{\linewidth}{!}
    {
    \begin{tabular}{lcccc}
    \toprule
        \textbf{Version} & \textbf{w/ Feedback} & \textbf{w/o Feedback} & $\mathbf{\Delta_{C\to I}}$ & $\mathbf{\Delta_{I\to C}}$  \\ \midrule
         3.5-Turbo & 36.75 & 21.37 & 20.51 & 5.13 \\
         4o-mini & 38.53 & 11.93 & 29.36 & 2.75 \\
         4-Turbo & 29.31 & 12.07 & 17.24 & 0.00 \\
         4o & 47.87 & 24.47 & 25.53 & 2.13 \\
         \bottomrule
    \end{tabular}
    }
    \caption{Answer correctness of the GPT family of models in queries containing invocation error. `w/ Feedback' and `w/o Feedback' represent cases where detailed feedback or only simple error reporting is provided, respectively. `$\mathbf{\Delta_{C\to I}}$' denotes the proportion of correct answers that become incorrect, while `$\mathbf{\Delta_{I\to C}}$' represents the proportion of incorrect answers that become correct, when transitioning from detailed feedback to simple error reporting.}
    \label{tab:feedback}
\end{table}

\section{Related Works}
\paragraph{LLMs in Tool Use}

The use of tools is a prominent hallmark of biological intelligence~\cite{tool-human}. Equipping LLMs with the ability to use tools is therefore a key milestone in advancing their capabilities toward artificial general intelligence~\cite{analy-ye, agent-survey}. Tools broadly encompass APIs, online services, application software, and other models that can be represented in formats accessible to LLMs~\cite{tool-learning}.
A critical factor in enhancing tool-use performance is constructing extensive datasets that detail tool use~\cite{Toolalpaca, ToolACE}. This involves generating diverse user queries and their corresponding tool sets. Existing approaches often employ a tool-driven methodology, collecting tools from various sources and using models to simulate user queries~\cite{ToolQA, steptool}. However, these methods lack diversity, fail to ensure dependency consistency, and cannot reliably verify data correctness.
In this paper, we propose a query-driven data construction approach. This method extends the range of locally executable tools through multi-hop queries, improving dataset quality and better supporting the development of LLM tool-use capabilities.

\paragraph{Evaluation of Tool Use}

Effectively evaluating the tool-use capabilities of LLMs is crucial for identifying their strengths and weaknesses. Existing methods, such as manual verification~\cite{Toolalpaca} or checking for the presence of a final answer~\cite{Toolllm}, fall short in providing objective and reliable measures of performance. Multi-dimensional approaches~\cite{ToolEyes, ToolSword} attempt to evaluate the process and outcomes of tool use but risk introducing model bias and inconsistencies.
In this paper, we focus on evaluating LLMs in multi-hop tool use scenarios. Our query-driven data construction scheme predefines verifiable answers, ensuring accurate assessments and providing a robust framework for evaluation.

\section{Conclusion}

In this paper, we introduce ToolHop, a novel dataset designed to evaluate LLMs in multi-hop tool use. ToolHop employs a query-driven data construction framework, encompassing tool creation, document refinement, and code generation. This approach overcomes the limitations of previous methods, ensuring diverse queries, meaningful interdependencies, locally executable tools, detailed feedback, and verifiable answers.
Using ToolHop, we benchmark 14 LLMs across five families, providing a comprehensive evaluation of their tool-use capabilities. Further studies illuminate the distinct characteristics of different LLM families, offering actionable insights to enhance their performance. By setting a robust standard for multi-hop tool use evaluation, ToolHop lays the groundwork for advancing LLMs' ability to perform complex tool-based reasoning tasks.

\section*{Limitations}

While our dataset effectively evaluates the performance of LLMs in multi-hop tool use, one limitation of this work is the lack of an immediate strategy for enhancing these capabilities. Nonetheless, the scalability of our data construction scheme represents a significant advantage, as it can be readily adapted to create training datasets aimed at addressing this challenge. We hypothesize that targeted training using such datasets could markedly improve the ability of LLMs to perform multi-hop tool use tasks.
Additionally, we provide a detailed analysis of current tool-use characteristics in LLMs, offering valuable insights that can serve as a foundation for future research and advancements in this area.

\section*{Acknowledgments}
The authors wish to thank the anonymous reviewers for their helpful comments. This work was partially funded by National Natural Science Foundation of China (No. 62476061,62206057,62076069), Shanghai Rising-Star Program (23QA1400200), Natural Science Foundation of Shanghai (23ZR1403500), Program of Shanghai Academic Research Leader under grant 22XD1401100.

\bibliography{custom}

\newpage
\clearpage

\appendix
\onecolumn

\section{Comparison of Various Benchmarks}
\label{sec:comparison}

\begin{table}[H]
    \centering
    \resizebox{\linewidth}{!}
    {
    \begin{tabular}{lccccccc}
    \toprule
       \multirow{2}*{\textbf{Benchmarks}} &  \multirow{2}*{\textbf{\# Tools}} &  \multirow{2}*{\textbf{\# Instances}} &  \multirow{2}*{\textbf{Multi-Tool?}} &  \textbf{Meaningful} & \textbf{Locally} & \textbf{Detailed} & \textbf{Verifiable} \\
       & & & &  \textbf{Interdependencies?} &  \textbf{Executable Tools?} &  \textbf{Feedback?} &  \textbf{Answers?} \\
\midrule
API-Bank~\cite{APIBank} & 73 & 314 & \textcolor{green!50!black}{\checkmark} & \textcolor{red}{\texttimes} & \textcolor{green!50!black}{\checkmark} & \textcolor{green!50!black}{\checkmark} & \textcolor{green!50!black}{\checkmark} \\
ToolAlpaca~\cite{Toolalpaca} & 426 & 3938 & \textcolor{red}{\texttimes} & \textcolor{red}{\texttimes} & \textcolor{red}{\texttimes} & \textcolor{red}{\texttimes} & \textcolor{red}{\texttimes} \\
RestBench~\cite{RestGPT} & 94 & 157 & \textcolor{green!50!black}{\checkmark} & \textcolor{red}{\texttimes} & \textcolor{red}{\texttimes} & \textcolor{red}{\texttimes} & \textcolor{green!50!black}{\checkmark} \\
ToolBench~\cite{Toolllm} & 16464 & 126484 & \textcolor{green!50!black}{\checkmark} & \textcolor{red}{\texttimes} & \textcolor{green!50!black}{\checkmark} & \textcolor{green!50!black}{\checkmark} & \textcolor{red}{\texttimes} \\
MetaTool~\cite{metatool} & 199 & 21127 & \textcolor{green!50!black}{\checkmark} & \textcolor{red}{\texttimes} & \textcolor{red}{\texttimes} & \textcolor{red}{\texttimes} & \textcolor{green!50!black}{\checkmark} \\
TaskBench~\cite{TaskBench} & 103 & 23271 & \textcolor{green!50!black}{\checkmark} & \textcolor{red}{\texttimes} & \textcolor{green!50!black}{\checkmark} & \textcolor{green!50!black}{\checkmark} & \textcolor{red}{\texttimes} \\
T-Eval~\cite{TEval} & 15 & 533 & \textcolor{green!50!black}{\checkmark} & \textcolor{red}{\texttimes} & \textcolor{green!50!black}{\checkmark} & \textcolor{green!50!black}{\checkmark} & \textcolor{green!50!black}{\checkmark} \\
ToolEyes~\cite{ToolEyes} & 568 & 382 & \textcolor{green!50!black}{\checkmark} & \textcolor{red}{\texttimes} & \textcolor{green!50!black}{\checkmark} & \textcolor{green!50!black}{\checkmark} & \textcolor{red}{\texttimes} \\
UltraTool~\cite{Ultra} & 2032 & 5824 & \textcolor{green!50!black}{\checkmark} & \textcolor{red}{\texttimes} & \textcolor{red}{\texttimes} & \textcolor{red}{\texttimes} & \textcolor{red}{\texttimes} \\
Seal-Tools~\cite{Sealtools} & 4076 & 14076 & \textcolor{green!50!black}{\checkmark} & \textcolor{red}{\texttimes} & \textcolor{red}{\texttimes} & \textcolor{red}{\texttimes} & \textcolor{green!50!black}{\checkmark} \\
AnyToolBench~\cite{anytool} & - & 400 & \textcolor{green!50!black}{\checkmark} & \textcolor{red}{\texttimes} & \textcolor{green!50!black}{\checkmark} & \textcolor{green!50!black}{\checkmark} & \textcolor{red}{\texttimes} \\
BFCL v3 (multi-turn)~\cite{BFCL} & - & 1000 & \textcolor{green!50!black}{\checkmark} & \textcolor{green!50!black}{\checkmark} & \textcolor{red}{\texttimes} & \textcolor{red}{\texttimes} & \textcolor{green!50!black}{\checkmark} \\
ToolHop & 3912 & 995 & \textcolor{green!50!black}{\checkmark} & \textcolor{green!50!black}{\checkmark} & \textcolor{green!50!black}{\checkmark} & \textcolor{green!50!black}{\checkmark} & \textcolor{green!50!black}{\checkmark} \\
\bottomrule
    \end{tabular}
    }
    \caption{A comparison between ToolHop and existing benchmarks related to tool use.}
    \label{tab:comparison}
\end{table}

To illustrate the necessity and advantages of ToolHop, Table~\ref{tab:comparison} presents a comparison between ToolHop and existing benchmarks related to tool use. Most existing benchmarks (e.g., API-Bank) lack meaningful interdependencies between tools. This limitation prevents them from evaluating a model’s ability to perform multi-hop tool use, where the output of one tool serves as the input to another. The only benchmark that includes meaningful interdependencies, BFCL v3 (multi-turn), lacks locally executable tools and detailed feedback. Moreover, it only supports models that generate all tool calls at once, rather than allowing multi-turn interactions with the environment. This makes it unsuitable for evaluating dynamic tool use, such as reasoning updates based on feedback.

In contrast, ToolHop is the only benchmark that incorporates all five key features: diverse queries, meaningful interdependencies, locally executable tools, detailed feedback, and verifiable answers. These features make it a comprehensive and realistic dataset for evaluating model performance in multi-hop tool use scenarios.

Additionally, most existing datasets become obsolete within 1–2 months as models quickly surpass their challenges. However, even the most advanced models continue to struggle with ToolHop. To maintain its effectiveness over time, we actively expand the dataset with more difficult tasks, establishing it as a long-term and instructive benchmark for evaluating tool-use capabilities.

\clearpage

\section{Prompt for Data Construction}
\label{sec:prompt-data}

Our proposed query-driven data construction scheme involves tool creation, document refinement, and code generation. The prompts used for each process are provided in Table~\ref{tab:prompt-tool-creation}, Table~\ref{tab:prompt-document-refinement}, and Table~\ref{tab:prompt-code-generation}, respectively.

\begin{table}[H]
    \centering
    \resizebox{\linewidth}{!}
    {
    \begin{tabular}{m{\linewidth}}
    \toprule
    Identify the appropriate tool to solve the given problem and provide an analysis of the tool design. The output should be in JSON format, following the specified structure.\\
\\
\# Steps\\
\\
1. **Analyze the Problem**: Understand the question and determine the type of information required to answer it.\\
2. **Tool Design**: Design a tool that can solve the problem, considering the complexity and additional functionalities it might need.\\
3. **Parameter Specification**: Define the parameters for the tool, ensuring they are comprehensive and flexible for various use cases.\\
4. **Output Construction**: Format the output in JSON, including both the analysis and the tool schema.\\
\\
\# Notes\\
\\
- Ensure the tool is versatile enough to handle similar queries for different sports figures.\\
- Consider edge cases.\\
\\
\# Output Format\\
\\
The output should be a JSON object with the following structure **without any other contents**:\\
- "analysis": A detailed analysis of the ideas behind the tool design.\\
- "tool": A JSON schema characterizing the tool, including its name, description, and parameters.\\
\\
\# Example\\
\\
\{Example\}\\
\\
**Question**: \{Question\}\\
\\
**Output**:\\
\\
\bottomrule
    \end{tabular}
    }
    \caption{The prompt for tool creation, where `\{Example\}' and `\{Question\}' represent the example and subquery, respectively.}
    \label{tab:prompt-tool-creation}
\end{table}

\clearpage

\begin{table}[H]
    \centering
    \resizebox{\linewidth}{!}
    {
    \begin{tabular}{m{\linewidth}}
    \toprule
    Refine the design of a tool by enhancing its description and increasing the complexity of parameters (e.g., numbers and types) while maintaining compatibility with the original functionality.\\
\\
\# Steps\\
\\
1. **Analyze the Current Tool**: Examine the existing tool's description and parameters to understand its functionality and limitations.\\
2. **Identify Areas for Refinement**: Determine which aspects of the tool can be improved or expanded to better meet real-world requirements.\\
3. **Refine the Description**: Enhance the tool's description to clearly articulate its refined functionality.\\
4. **Add and Refine Parameters**: Introduce new parameters or refine existing ones to increase complexity and utility, ensuring they align with the original functionality.\\
5. **Ensure Compatibility**: Verify that the refined version remains compatible with the original tool's purpose and structure.\\
\\
\# Output Format\\
\\
The output should be in JSON format with the following structure **without any other contents**:\\
\{\\
  "analysis": "Analysis of ideas about refining the tool.",\\
  "refined\_version": the version after refinement, should be follow JSON SCHEMA format as the original tool\\
\}\\
\\
\# Notes\\
\\
- Ensure that any new parameters added are relevant and enhance the tool's functionality.\\
- Maintain backward compatibility with the original tool's design and purpose.\\
\\
**Tool**:\\
\{Tool\}\\
\bottomrule
    \end{tabular}
    }
    \caption{The prompt for document refinement, where `\{Tool\}' represents the preliminary document.}
    \label{tab:prompt-document-refinement}
\end{table}

\clearpage

\begin{table}[H]
    \centering
    \resizebox{\linewidth}{!}
    {
    \begin{tabular}{m{\linewidth}}
    \toprule
    Create a function implementation based on a provided tool document, question, and answer. The function should strictly adhere to the tool's specifications, including the function name, parameter names, and types. Ensure the function is fully realized and capable of returning different feedback based on the input parameters.\\
\\
\# Steps\\
\\
1. **Understand the Tool Document**: Review the tool document to identify the function name, parameter names, and types.\\
2. **Analyze the Question and Answer**: Determine how the function should be used to answer the question.\\
3. **Implement the Function**:\\
   - Use the tool name as the function name.\\
   - Define parameters exactly as specified in the tool document.\\
   - Implement the function logic to produce the correct answer for the given question.\\
   - Simulate additional return values as specified in the tool document.\\
4. **Error Handling**: Develop a robust error handling mechanism to return valid error messages for incorrect inputs or other issues.\\
\\
\# Notes\\
\\
- Ensure parameter types and names match exactly with the tool document.\\
- Simulate additional return values as needed based on the tool's documentation.\\
- Implement comprehensive error handling to cover potential issues.\\
\\
\# Output format\\
\\
Output the result in JSON format with the following structure **without any other contents**:\\
\{
  "analysis": "Detailed analysis of how the function was designed, including reasoning for parameter choices and exception handling.",\\
  "function": "The specific function design, including code and comments explaining each part."\\
\}\\
\\
**Tool Document**:\\
\{document\}\\
\\
**Question**: \{question\}\\
\\
**Answer**: \{answer\}\\
\bottomrule
    \end{tabular}
    }
    \caption{The prompt for code generation, where `\{document\}', `\{question\}' and `\{answer\}' represent the refined document, the subquery and the corresponding answer, respectively.}
    \label{tab:prompt-code-generation}
\end{table}

\clearpage
\section{Prompt for Domain Classification}
\label{sec:prompt-domain}

We conduct a domain analysis of the queries in ToolHop using GPT-4o, with the corresponding prompts provided in Table~\ref{tab:prompt-domain}.

\begin{table}[H]
    \centering
    \resizebox{\linewidth}{!}
    {
    \begin{tabular}{m{\linewidth}}
    \toprule
    Identify the domain of the given sentence by analyzing its content and context. The domain should be a single, specific category that best describes the subject matter of the sentence.\\
\\
\# Steps\\
\\
1. **Analyze the Sentence**: Break down the sentence to understand its components and context.\\
2. **Identify Key Elements**: Look for specific terms or phrases that indicate the subject matter, such as names, dates, or specific topics.\\
3. **Determine the Domain**: Based on the analysis, select the most appropriate domain that encapsulates the main focus of the sentence.\\
\\
\# Output Format\\
\\
The output should be in JSON format with the following structure **without any other contents**:\\
\{\\
  "analysis": "Analysis of the given sentence.",\\
  "domain": The domain of the sentence, as short as possible\\
\}\\
\\
\# Notes\\
\\
- Ensure the domain is specific and directly related to the main subject of the sentence.\\
- Consider the broader context if the sentence includes specific names or events.\\
\\
Sentence: \{sentence\}\\
    \\
    \bottomrule
    \end{tabular}
    }
    \caption{The prompt for domain classification, where `\{sentence\}' represents the multi-hop query.}
    \label{tab:prompt-domain}
\end{table}

\clearpage

\section{Details for Models}
\label{sec:detail-model}

We evaluate 14 LLMs from five families, spanning both open- and closed-source models, to provide a comprehensive analysis of their performance in multi-hop tool use.  

\begin{itemize}  
    \item \textbf{LLaMA3.1 Family.} The LLaMA3.1 family, developed by Meta, includes open-source LLMs with model sizes of 8B, 70B, and 405B, and context lengths up to 128K. These models are optimized for tasks such as long text summarization, multilingual dialogue, and code generation. Due to computational constraints, this study evaluates \textbf{LLaMA3.1-Instruct-8B} and \textbf{LLaMA3.1-Instruct-70B}.  

    \item \textbf{Qwen2.5 Family.} The Qwen2.5 family, developed by Alibaba, consists of open-source LLMs pre-trained on 18 trillion tokens. These models are designed to excel in mathematics, programming, and knowledge representation, with versions ranging from 0.5B to 72B. Our evaluation focuses on \textbf{Qwen2.5-Instruct-7B}, \textbf{Qwen2.5-Instruct-14B}, \textbf{Qwen2.5-Instruct-32B}, and \textbf{Qwen2.5-Instruct-72B}.  

    \item \textbf{Gemini1.5 Family.} The Gemini1.5 family, developed by DeepMind, utilizes a mixture-of-experts~\cite{MoE} architecture for advanced reasoning across large datasets. This family includes flash and pro versions. For this paper, we analyze \textbf{Gemini1.5-flash-002} and \textbf{Gemini1.5-pro-002}.  

    \item \textbf{Claude3.5 Family.} The Claude3.5 family, developed by Anthropic, includes closed-source Haiku and Sonnet versions, which are known for advancements in instruction-following and nuanced reasoning. This evaluation considers \textbf{Claude3.5-Haiku} and \textbf{Claude3.5-Sonnet}.  

    \item \textbf{GPT Family.} The GPT family, developed by OpenAI, comprises closed-source LLMs designed for text generation, multimodal understanding, and tool use. In this paper, we evaluate \textbf{GPT-3.5-Turbo}, \textbf{GPT-4o-mini}, \textbf{GPT-4-Turbo}, and \textbf{GPT-4o}.  
\end{itemize}  

\clearpage

\section{Performance under Various Inference Frameworks}
\label{sec:further}

\begin{table}[H]
    \centering
    {
    \begin{tabular}{lllccccc}
    \toprule
    {\textbf{Source}} & {\textbf{Family}} & {\textbf{Version}} & \textbf{FC} & \textbf{ReAct} & \textbf{CodeAct} \\ \midrule
    \multirow{2}*{Tool-Use} & ToolLLaMA-2 & 7B-v2 & - & 16.08  & - \\ \cmidrule{2-6}
    & TL-CodeLLaMA-2 & 7B & - & - & 17.59 \\ \midrule
      \multirow{6}*{Open-Source} & \multirow{2}*{LLaMA3.1} & Instruct-8B & 13.47 & 38.59 & 25.63 \\
      & &Instruct-70B & 12.76 & 57.79 & 44.19 \\ \cmidrule{2-6}
      & \multirow{4}*{Qwen2.5} & Instruct-7B & 16.18 & 40.50 & 22.91 \\
      & & Instruct-14B & 26.13 & 46.13 & 36.78 \\
      & & Instruct-32B & 22.61 & 48.04 & 45.73 \\
      & & Instruct-72B & 38.29 & 48.44 & 46.63 \\ \midrule
      \multirow{8}*{Closed-Source} & \multirow{2}*{Gemini1.5} & flash-002 & 32.76 & 4.82 & 5.13 \\
      & & pro-002 & 33.07 & 22.61 & 7.24 \\ \cmidrule{2-6}
      & \multirow{2}*{Claude3.5} & Haiku & 44.72 & 34.07 & 48.34 \\
      & & Sonnet & 45.23 & 44.52 & 49.15 \\ \cmidrule{2-6}
      & \multirow{4}*{GPT} & 3.5-Turbo & 36.58 & 28.14 & 25.63 \\
      & & 4o-mini & 43.42 & 47.14 & 40.80 \\
      & & 4-Turbo & 46.83 & 53.67 & 42.61 \\
      & & 4o & 47.74 & 54.17 & 44.92 \\
      \bottomrule
    \end{tabular}
    }
    \caption{Performance of various LLMs on ToolHop across different inference frameworks. `FC' refers to answer correctness in the Free scenario using built-in function-calling templates.}
    \label{tab:result_temp}
\end{table}

In addition to using the built-in chat template, we also compare the performance of the models under other commonly used tool-use frameworks. Table~\ref{tab:result_temp} shows how different models perform with the ReAct~\cite{React} and CodeAct~\cite{CodeAct} inference frameworks, and compares these results to performance in the Free scenario. The results show that while model performance varies across frameworks, none of them effectively solves the challenges presented by ToolHop in any setting, which further supports our conclusions.

We also observe the following:
\begin{itemize}
    \item Open-source models benefit more from the ReAct framework, as it enforces a step-by-step reasoning process before tool invocation, which improves planning ability. In contrast, models using FC template often invoke tools directly without explicit reasoning, suggesting that open-source models still have room to improve their built-in function-calling capabilities.
    \item Closed-source models are generally well-optimized for FC, and their performance remains more stable across different frameworks compared to open-source models. However, different models still show preferences for specific frameworks. For instance, the GPT series performs best with ReAct, while the Claude 3.5 series prefers CodeAct, likely due to differences in training data and methodology.
\end{itemize}

\clearpage
\section{Case Study}
\label{sec:example}

\begin{figure}[H]
    \centering
    \includegraphics[width=0.9\linewidth]{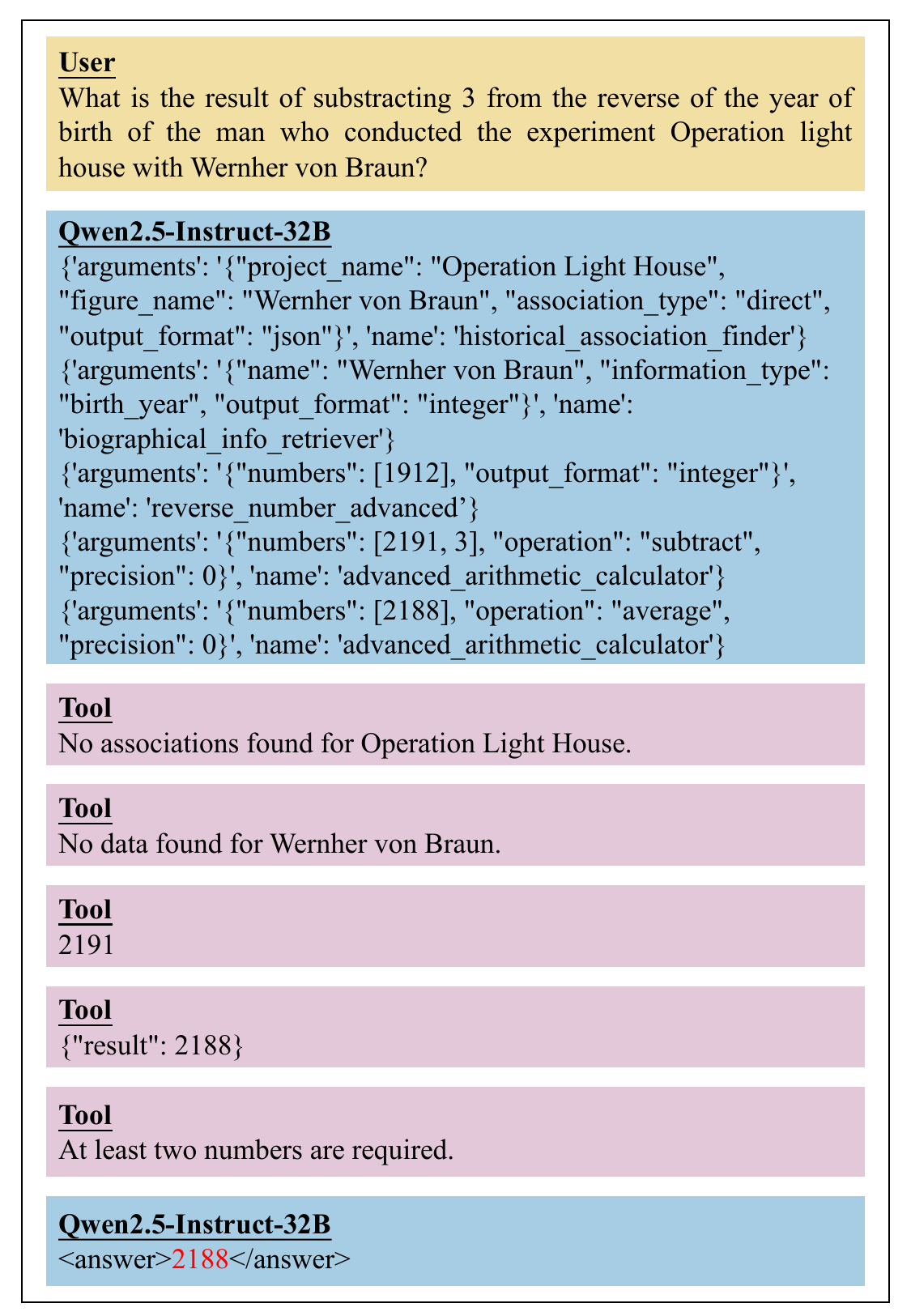}
    \caption{The Qwen2.5 family of LLMs emphasizes parallel tool calls in the mandatory tool use scenario, which can lead to hallucinations and incorrect answers.}
    \label{fig:example-qwen}
\end{figure}

\begin{figure}[H]
    \centering
    \includegraphics[width=0.9\linewidth]{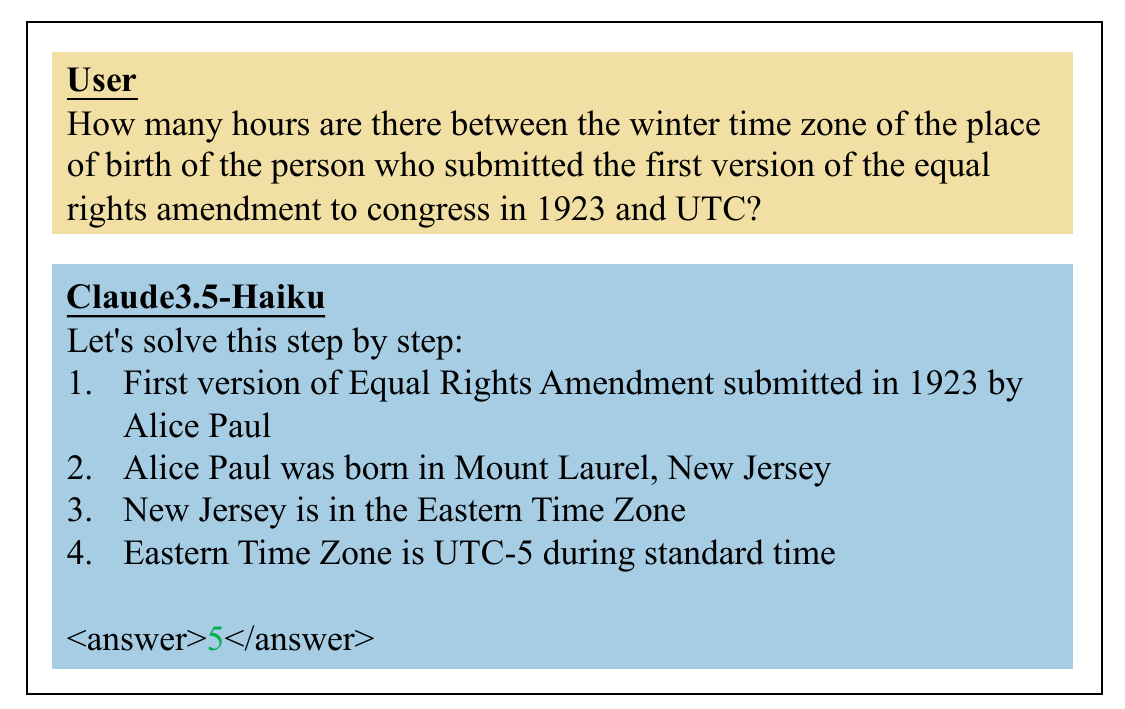}
    \caption{The Claude 3.5 family of LLMs optimizes CoT reasoning in the direct answer scenario, enhancing their analytical and problem-solving capabilities."}
    \label{fig:example-claude}
\end{figure}

\begin{figure}[H]
    \centering
    \includegraphics[width=0.9\linewidth]{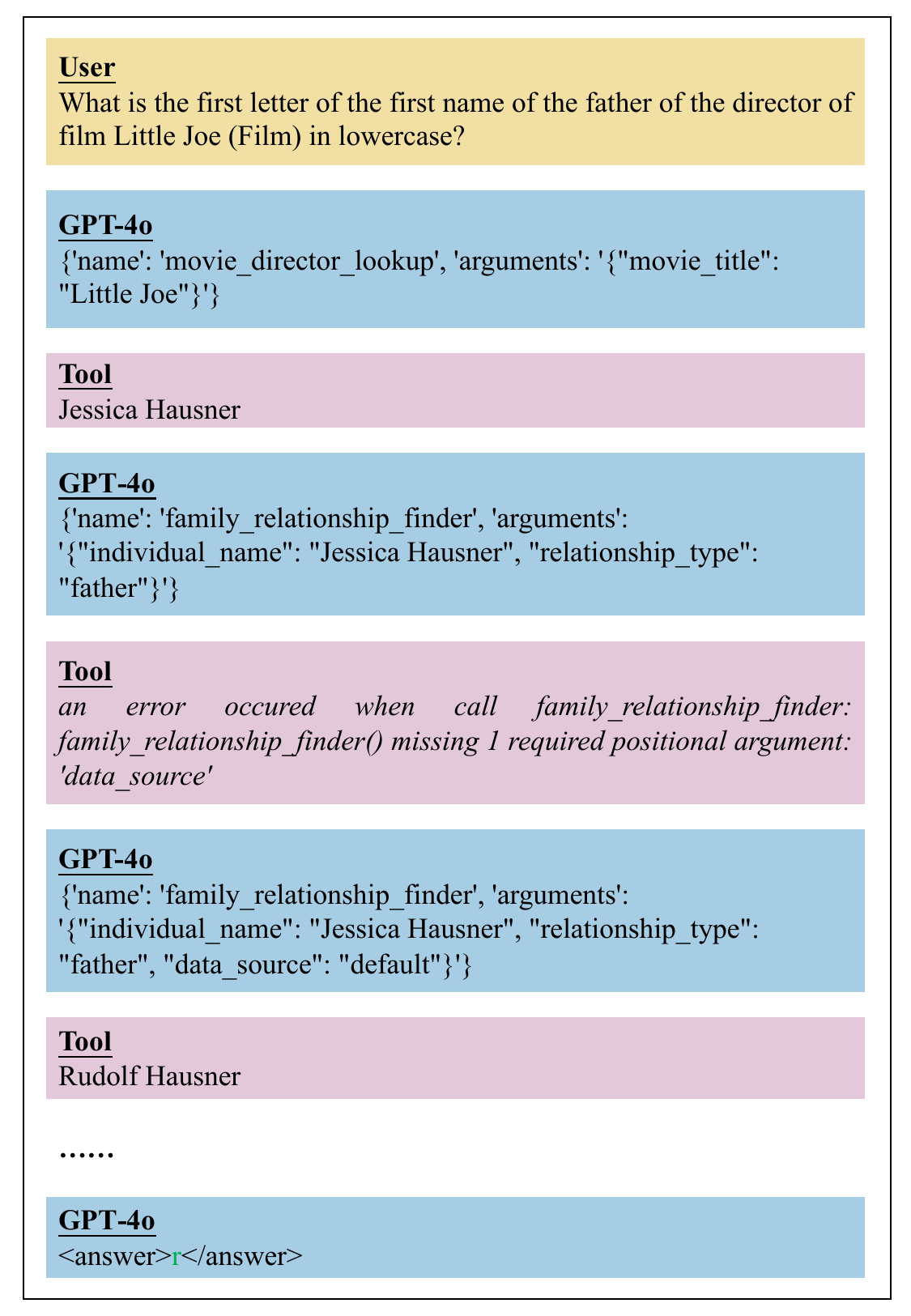}
    \caption{The GPT family of LLMs improves performance by refining calling behavior through the use of detailed tool feedback.}
    \label{fig:example-gpt}
\end{figure}

\begin{figure}[H]
    \centering
    \includegraphics[width=0.9\linewidth]{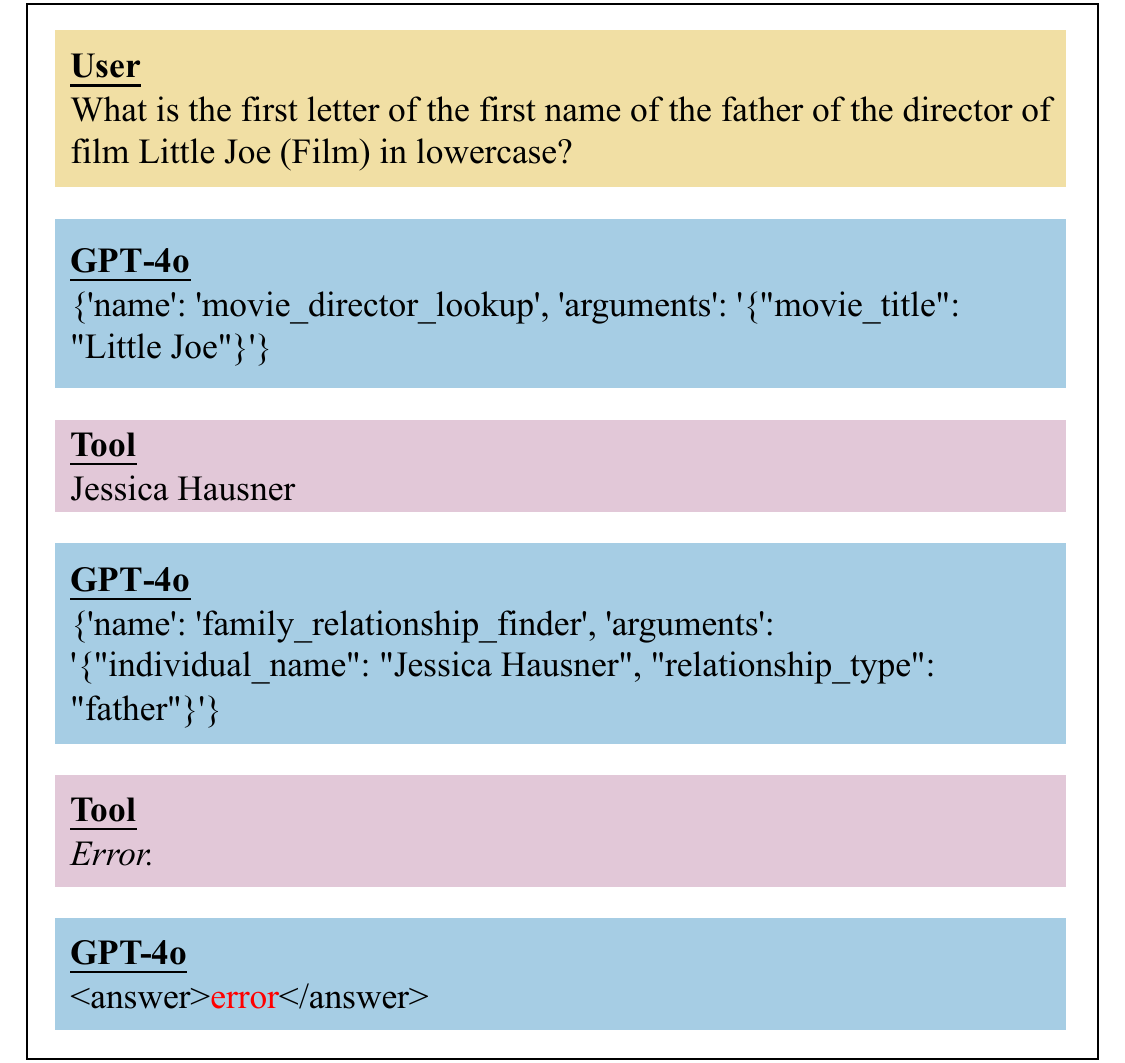}
    \caption{The GPT fmaily of LLMs struggles to correct their calling behavior when provided with minimal feedback.}
    \label{fig:example-gpt-nof}
\end{figure}

In Section~\ref{sec:analysis} and Section~\ref{sec:case}, we analyze the performance of different LLMs across various scenarios. In this section, we present relevant examples from Figure~\ref{fig:example-qwen} to Figure~\ref{fig:example-gpt-nof}.

\clearpage

\clearpage

\clearpage

\clearpage
\section{Examples of Tool Documents}
\label{sec:example-document}

Our query-driven data construction scheme generates preliminary document prior to refinement. Below, we provide examples of documents before and after refinement. As shown, the refinement process enhances the tool's functionality, increases the number of parameters and introduces more diverse parameter types.

\begin{verbatim}
{
    "name": "album_release_date_finder",
    "description": "A tool designed to find the release date of music
    albums. It queries a database or API to retrieve accurate 
    information about album release dates, accommodating variations in 
    album titles and artist names.",
    "parameters": {
        "type": "object",
        "properties": {
            "album_name": {
                "type": "string",
                "description": "The name of the album for which the 
                release date is being queried."
            },
            "artist_name": {
                "type": "string",
                "description": "The name of the artist or band 
                associated with the album, to ensure accuracy in case 
                of albums with similar names."
            },
            "output_format": {
                "type": "string",
                "enum": [
                    "date",
                    "text"
                ],
                "description": "The format of the output. Defaults to 
                date (the release date in YYYY-MM-DD format)."
            }
        },
        "required": [
            "album_name"
        ]
    }
}
    
\end{verbatim}

\begin{verbatim}
{
    "name": "album_release_date_finder",
    "description": "An advanced tool designed to find the release date 
    of music albums. It queries a comprehensive database or API to 
    retrieve accurate information about album release dates, 
    accommodating variations in album titles, artist names, album 
    versions, release regions, and languages. This tool ensures 
    precision and flexibility in retrieving album release information.",
    "parameters": {
        "type": "object",
        "properties": {
            "album_name": {
                "type": "string",
                "description": "The name of the album for which the 
                release date is being queried."
            },
            "artist_name": {
                "type": "string",
                "description": "The name of the artist or band 
                associated with the album, to ensure accuracy in case
                of albums with similar names."
            },
            "album_version": {
                "type": "string",
                "description": "The specific version of the album 
                (e.g., deluxe, remastered) to refine the search."
            },
            "release_region": {
                "type": "string",
                "description": "The geographical region where the album
                was released, which can affect the release date."
            },
            "language": {
                "type": "string",
                "description": "The language of the album, useful for
                albums released in multiple languages."
            },
            "output_format": {
                "type": "string",
                "enum": [
                    "date",
                    "text"
                ],
                "description": "The format of the output. Defaults to
                date (the release date in YYYY-MM-DD format)."
            }
        },
        "required": [
            "album_name"
        ]
    }
}
\end{verbatim}

\clearpage

\section{Examples of Code Implementations}
\label{sec:example-code}

Our query-driven data construction scheme translates the refined tool document into code, enabling it to function as a locally executable tool. Below, we provide the code implementation of the refined document in Appendix~\ref{sec:example-document}. The implementation fully realizes the defined functionality, provides valid feedback for correct parameter inputs, and incorporates a robust exception handling mechanism.

\begin{verbatim}
def album_release_date_finder(album_name: str, artist_name: str = 
    '', album_version: str = '', release_region: str = '', language: 
    str = '', output_format: str = 'date') -> str:
    """
    Finds the release date of a specified music album.

    Parameters:
    - album_name (str): The name of the album for which the release 
    date is being queried.
    - artist_name (str): The name of the artist or band associated with 
    the album.
    - album_version (str): The specific version of the album (e.g., 
    deluxe, remastered).
    - release_region (str): The geographical region where the album was 
    released.
    - language (str): The language of the album.
    - output_format (str): The format of the output, either 'date' or 
    'text'.

    Returns:
    - str: The release date of the album in the specified format.
    """
    # Simulated database/API response
    album_database = {
        'Boy': {
            'artist': 'U2',
            'release_date': '1980-10-20',
            'versions': {
                'standard': '1980-10-20',
                'deluxe': '2008-07-21'
            },
            'regions': {
                'US': '1980-10-20',
                'UK': '1980-10-20'
            },
            'languages': {
                'English': '1980-10-20'
            }
        }
    }

    # Error handling for required parameter
    if not album_name:
        return 'Error: The album_name parameter is required.'

    # Retrieve album information
    album_info = album_database.get(album_name)
    if not album_info:
        return 'Error: Album not found in the database.'

    # Check artist name if provided
    if artist_name and album_info['artist'] != artist_name:
        return 'Error: Artist name does not match the album record.'

    # Determine release date based on version, region, and language
    release_date = album_info['release_date']
    if album_version:
        release_date = album_info['versions'].get(album_version, 
        release_date)
    if release_region:
        release_date = album_info['regions'].get(release_region, 
        release_date)
    if language:
        release_date = album_info['languages'].get(language, 
        release_date)

    # Format the output
    if output_format == 'text':
        return f'The album "{album_name}" by {album_info["artist"]} was 
        released on {release_date}.'
    return release_date

# Example usage
print(album_release_date_finder(album_name='Boy', artist_name='U2', 
output_format='date'))  # Output: '1980-10-20'
print(album_release_date_finder(album_name='Boy', artist_name='U2', 
output_format='text'))  # Output: 'The album "Boy" by U2 was released 
on 1980-10-20.'

\end{verbatim}



\end{document}